\documentclass[conference]{IEEEtran}
\IEEEoverridecommandlockouts
\usepackage{cite}
\usepackage{amsmath,amssymb,amsfonts}
\usepackage{algorithmic}
\usepackage{graphicx}
\usepackage{textcomp}
\usepackage{hyperref}
\usepackage{xcolor}
\def\BibTeX{{\rm B\kern-.05em{\sc i\kern-.025em b}\kern-.08em
    T\kern-.1667em\lower.7ex\hbox{E}\kern-.125emX}}

\usepackage{bm}
\usepackage{comment}
\usepackage{subcaption}

\newcommand*{\ldblbrace}{\{\mskip-5mu\{}
\newcommand*{\rdblbrace}{\}\mskip-5mu\}}

\begin{document}


\title{BERTops: Studying BERT Representations under a Topological Lens}

\author{\IEEEauthorblockN{Jatin Chauhan, Manohar Kaul}
\IEEEauthorblockA{\textit{Dept. of Computer Science \& Engineering} \\
\textit{Indian Institute of Technology Hyderabad}\\
Email: \{chauhanjatin100, manohar.kaul\}@gmail.com}
}


\maketitle

\begin{abstract}
Proposing scoring functions to effectively understand, analyze and learn various properties of high dimensional hidden representations of large-scale transformer models like BERT can be a challenging task. In this work, we explore a new direction by studying the topological features of BERT hidden representations using persistent homology (PH). We propose a novel scoring function named ``persistence scoring function (PSF)" which: (i) accurately captures the homology of the high-dimensional hidden representations and correlates well with the test set accuracy of a wide range of datasets and outperforms existing scoring metrics, (ii) captures interesting post fine-tuning ``per-class`` level properties from both qualitative and quantitative viewpoints, (iii) is more stable to perturbations as compared to the baseline functions, which makes it a very robust proxy, and (iv) finally, also serves as a predictor of the attack success rates for a wide category of black-box and white-box adversarial attack methods. Our extensive correlation experiments demonstrate the practical utility of PSF on various NLP tasks relevant to BERT \footnote{Code is available \href{https://github.com/chauhanjatin10/BERTops}{here}. }.
\end{abstract}


\section{Introduction}
\label{sec:intro}
Extensive research is being conducted to comprehend the functionality of the \emph{transformer} \cite{vaswani2017attention} model and its variants \cite{devlin-etal-2019-bert} by studying them from various perspectives such as: \emph{language modeling}\cite{clark-etal-2019-bert}, \emph{generalization} \cite{hao2019visualizing}, \emph{robustness} and \emph{domain adaptation} standpoints. Many recent works have focused on proposing a wide range of scoring functions and metrics based on the attention mechanism of these models \cite{abnarquantifying, nEURIPS2019_159c1ffe} as well as their contextualized embeddings \cite{schmidt2020bert}, which can be used to assess the aforementioned properties \cite{elsahar_galle_annotate} and discover newer lines of work such as predicting their performance without explicitly training and/or testing \cite{xia_2020_predicting}. 

Scoring functions which can estimate and/or predict properties based on the hidden state representations of these models have diverse use cases. One such use case is the estimation of testing accuracies of the trained NLP models, which is helpful in the scenarios where held out sets are not available at all, some examples being online leaderboards \cite{wang2020superglue, rajpurkar2016squad}. One can argue that these held out test sets can be created by sampling from the training data itself or using techniques like cross-validation. However, this reduction in the size of the training set usually hurts the performance of deep learning models \cite{sun_2017_ICCV, effectiveness_4804817}. Another very interesting and highly practical use case is the estimation of the adversarial vulnerability of these fine-tuned BERT models, where we use the attack success rate of any given attacker (black or white-box) as the proxy for this experiment, without actually performing the attack. This benefits us greatly from a computational perspective, since performing an attack by developing a held-out sample set can be extremely resource-intensive as well as time-consuming. For example, we empirically observed that some black-box attacks can take nearly $15$ hours to generate adversarial examples for merely $1000$ sentences on a single dataset. We note that these estimations, however, also need to be reliable as well as invariant to slight input transformations, which can otherwise result in spurious and non-generalizable conclusions. One of the best ways to show it is by studying the stability of these scoring functions to perturbations in the space over which they are computed. 

In this work, we provide a novel scoring function based on the hidden state representations of the transformer variant BERT, using \textit{persistent homology (PH)}\cite{edelsbrunner2000topological, edelsbrunner2010computational}. We show that our scoring function, named \textbf{persistence scoring function (PSF)}, captures the \textit{$0$-th} and \textit{$1$-st} dimensional homology features which are essentially the \textit{connected components} and \textit{holes (or tunnels)} in the hidden representation space, and thus more accurately defines the spread of data points via their persistence values, which is otherwise difficult for functions based solely on Euclidean distance or dimensionality reduction techniques \cite{gracia2014}.

Through our empirical investigations, we first show that the scores generated by our proposed PSF correlate well to the test set accuracy of fine-tuned BERT models on a diverse category of datasets. We also verify PSF's superior scoring performance over strong existing baseline methods. More interestingly, we observed that PSF scores also shows high correlations to the f1-score per class label on the datasets and that the qualitative observations from this experiment can help in understanding some properties of the fine-tuned models which are retained from the pre-training phase. We then show that PSF exhibits stability against noisy perturbations in the input space and thus serves as a robust scoring function. Finally, based on the aforementioned preliminary experiments, we show that PSF can also serve as an estimator of vulnerability against both black-box and white-box adversarial attacks and significantly outperforms the state-of-the-art vulnerability scoring baselines. 

We also note the parallel between PSF and the problem of predicting the \textit{generalization gap (GP)} \cite{jiang2019predicting}, which is the difference between a model’s performance on the training data and its performance on the unseen data drawn from the same distribution, for which various bounds based on \emph{complexity measures} such as the VC dimension and Rademacher complexity have been proposed to ``explain and interpret" how sensitive a neural network’s output is to changes in the input space (e.g., added noise, distribution shifts etc). The rationale behind PSF follows similar lines, but more importantly, it is \emph{invariant to slight input transformations} and also simultaneously serves as a strong indicator of a network’s likelihood of being successfully attacked, for which no prior work in NLP exists. It is worth noting that in the absence of such complexity measures and estimators, one has to resort to generating numerous \emph{augmented training sets} along with their test sets for all possible input transformations (including adversarial examples), which can be computationally infeasible.

Our formulation of PSF also has a very direct connection to the compactness of Euclidean embeddings of the data points (the work of \cite{DBLP:journals/corr/abs-0712-2638} provides relevant bounds for the same). 
Intuitively speaking, lower the PSF score for a given cluster of points in Euclidean space, more compact the cluster is, which ultimately accounts for well separated classes and improved performance (section \ref{subsec:results_}). Subsequently, higher PSF correlates to higher adversarial vulnerability, since the intra-cluster spread is larger, making the model prone to attacks, as discussed in more detail in section \ref{sec:adv_vul}.


The summary of our contributions is as follows: 
\begin{enumerate}
    \item To the best of our knowledge, we are the first to propose a scoring function (i.e., PSF) using topological features of BERT's hidden representation space.
    \item Extensive experiments demonstrate that PSF can be used as an estimator of important properties of fine-tuned BERT models and is stable against perturbations.
    \item Our method requires only a few lines of post-processing code and can be utilized with any representation learning method, in and even outside the realm of NLP models.
\end{enumerate}

\section{Related Work}
\label{sec:related_work}
Using scoring functions as estimators and/or predictors of some property of deep learning models and datasets has gained widespread interest in recent years. \cite{dataset_ot} used optimal transport distance between datasets to predict the performance of BERT on \emph{domain adaptation} from one dataset to another, \cite{nEURIPS2019_faf02b23} worked on word embedding matrix compression algorithms and introduced an \textit{eigen overlap score} to identify the best performing compression method, without actually training downstream models on them. \cite{xia_2020_predicting} tried to predict the evaluation score of an NLP experiment for machine translation, while the work of \cite{elsahar_galle_annotate} tries to predict the performance drop of modern NLP models under domain-shift. 

Complementary to other mathematical sub-fields, \emph{persistent homology} (PH) has also been used for such works. \cite{rieck2018neural} proposed a complexity measure, named \emph{neural persistence}, based on PH to characterize the structural complexity of MLPs, whereas\cite{corneanu_9156398, corneanu_8953424} proposed PH-based algorithms to predict the ability of models to generalize and detect adversarial samples. 
Both these PH-based methods employ methods that are fundamentally different than ours, namely (i) their filtrations are constructed on binary graphs as opposed to our filtration construction which are built on BERT hidden representations, and (ii) their methods mainly focus on computer vision related tasks. 

PH has also been previously used in NLP by \cite{zhu2013persistent, doshi2018using, kushnareva-etal-2021-artificial} for document analysis, standard classification tasks and studying BERT's attention matrix, 
however, none of these prior works have studied the topological features of the space of BERT hidden representations via PH.

\begin{figure*}[tbp]
	\begin{subfigure}[b]{0.25\textwidth}
		\centering
		\includegraphics[width=\linewidth]{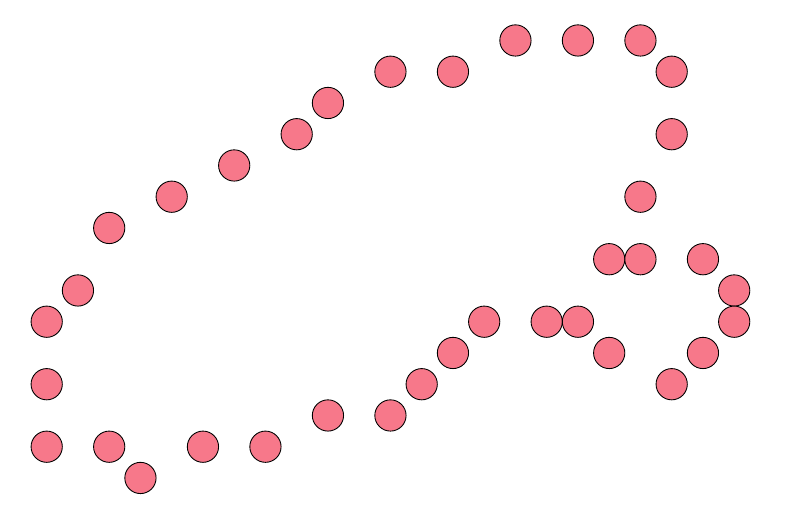}
		\caption{$X=X_0$}
		\label{fig:pd1}
	\end{subfigure}%
	\begin{subfigure}[b]{0.25\textwidth}
		\centering
		\includegraphics[width=\linewidth]{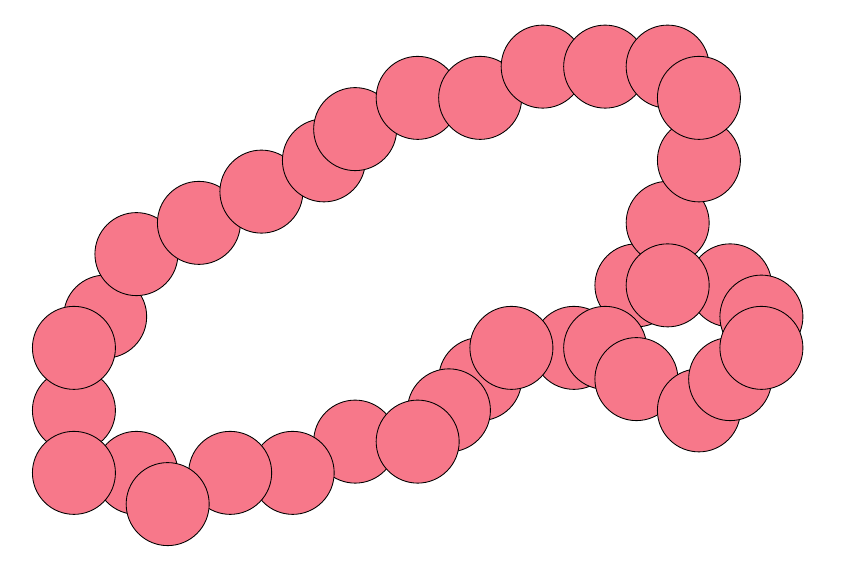}
		\caption{ $X_{0.2}$ }
		\label{fig:pd2}
	\end{subfigure}%
	\begin{subfigure}[b]{0.25\textwidth}
		\centering
		\includegraphics[width=\linewidth]{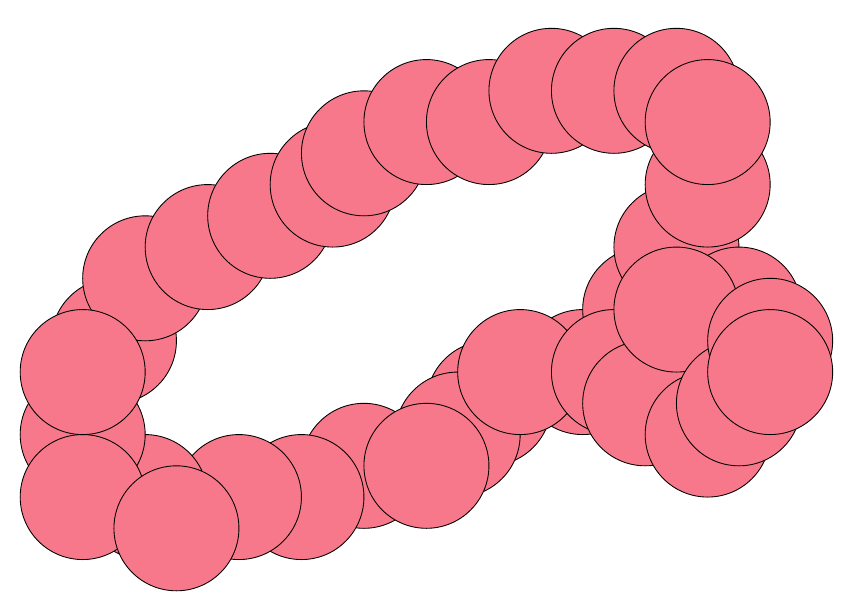}
		\caption{$X_{0.35}$ }
		\label{fig:pd3}
	\end{subfigure}%
	\begin{subfigure}[b]{0.25\textwidth}
		\centering
		\includegraphics[width=\linewidth]{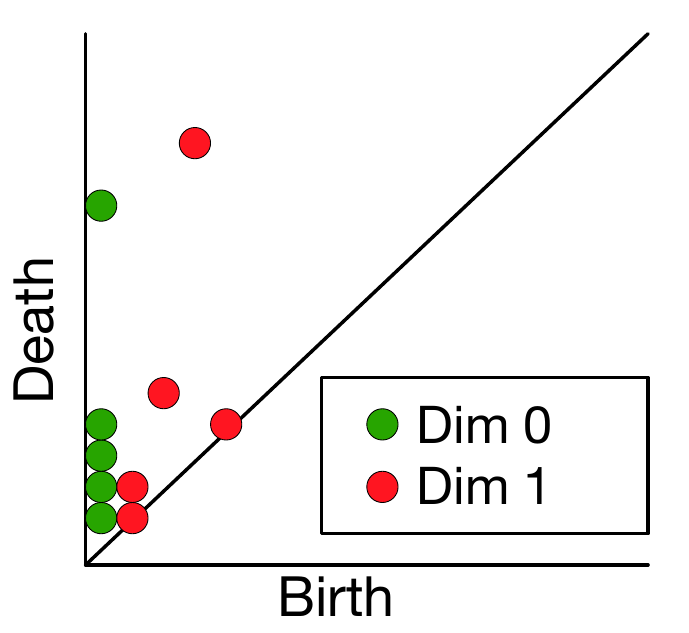}
		\caption{Persistence diagram}
		\label{fig:pd4}
	\end{subfigure}
	\caption{ 
		(a) Original data points $X$ with $r=0$. (b,c) ``Fattened points" by placing balls of radius $r >0$ centered on each point
		in $X$ and generating spaces $X_{0.2}$ and $X_{0.35}$. As we increase $r$, one of the topological 
		features (tunnels) born at $r=0.2$, dies at $r=0.35$. (d) The persistence diagram summarizes the lifetimes of the topological features of $X$. The green dots represent the connected components and the red dots represent the tunnels formed.
}	
	\label{fig:pers}
\end{figure*}

\section{Background}
\label{sec:background}
\textbf{Persistent Homology (PH)}:
In this section, we briefly review \emph{persistent homology} (PH). For a more complete coverage of algebraic topology and PH, we refer the reader to\cite{edelsbrunner2010computational}.

Let $X = \{x_1, \cdots , x_n  \} $ be a compact subset in a metric space $(M,d)$, where $d_M: M \rightarrow \mathbb{R}$ is the underlying distance function. Consider a ball $B_r(x) = \{  y \in M \mid d_M(x,y) \leq r  \} $ of radius $r$ centered on $x \in X$.
Imagine placing such a ball of radius $r$ on every point in $X$ to get a model 
$X_r = \cup_{i=1}^n B_r(x_i)$; note that $X_0=X$.

As we vary the radius $r$ from $0$ to $\infty$, (as shown in Figures~\ref{fig:pd1}--\ref{fig:pd3} ), we observe changes in $X_r$.
Persistent homology summarizes the change in topology of $X_r$, as $r$ is increased. Here, $X_r$ is considered as a topological 
space and its $j$-th homology group $H_j(X_r)$ ($j=0,1,\cdots)$, is a vector space, and its dimension $\dim H_j(X_r)$ depicts 
the \emph{number of connected components} ($j=0$), \emph{tunnels/holes} ($j=1$), \emph{voids} ($j=2$), and so on.
For example, $X_{0.2}$ in Figure~\ref{fig:pd2} comprises of one connected component (i.e., $\dim H_0(X_{0.2} )  = 1$) 
and two tunnels (i.e., $\dim H_1(X_{0.2} )  = 2$).

Notice that these topological features can \emph{be born} (appear) and \emph{die} (disappear) as the radius $r$ is increased.
For example, one of the tunnels in $X_{0.2}$ dies in $X_{0.35}$.
By gathering the \emph{birth-death} pairs $P_l(X) = \ldblbrace  (b_i,d_i) \in \mathbb{R}^2 \mid i \in I \rdblbrace$ , we obtain a multi-set. The collection $P_l(X)$ refers to the $l$-th persistence diagram, where $l$ is the dimension of the homology group. 
Figure~\ref{fig:pd4} shows the birth-death pairs corresponding to the number of connected components (in green) and the tunnels (in red) in the same persistence diagram. Points close to the diagonal are topological features with short lifespans and are considered \emph{topological noise}. We are interested in features that \emph{persist} over larger intervals of $r$.

\section{Persistence Scoring Function (PSF)}
\label{sec:PSF}
Here, we propose a novel scoring function called \emph{persistent scoring function (PSF)}. Let $X$ denote the finite and compact subset of BERT hidden state vectors. As explained in Section~\ref{sec:background}, we can study the topological features, expressed at various spatial scales, of the BERT hidden representation space (i.e., $X$). $X$ is studied as a collection of $l$ persistence diagrams and denoted by $P_l(X)$. 

Inspired by\cite{gabrielsson20a}, we propose our PSF. First, we concatenate the collection of persistence diagrams, denoted by $P(X)$, as 
\[
P(X) = \bigoplus_{l=0}^{l_m} P_l(X)
\] 
where $\bigoplus$ denotes concatenation of persistence diagrams represented as multi-sets of birth-death pairs into a single multi-set $P(X)$ and $l_m$ is the maximum dimension of the computed homology group. This concatenation operation is different from the \emph{union of multi-sets} and preserves the total multiplicity of birth-death pairs across all persistence diagrams. 

Our PSF based on the concatenated persistence diagrams $P(X)$, denoted by $\mathcal{L}(p,q; P(X))$, is then computed as
\begin{align}
	\label{eq:pers}
 \frac{1}{|P(X)| } \sum_{i=1}^{ |P(X)| }   
  \left( \frac{| d_i - b_i  |}{\hat{d}} \right) ^p 
\left( \frac{| d_i + b_i  |}{2 \hat{d}} \right)^q
\end{align}
where $|P(X)|$ is the cardinality of the concatenated multiset and $\hat{d}$ is computed as: $max\{d_i | (b_i, d_i) \in P(X)\}$.
In \eqref{eq:pers}, the term $|d_i - b_i|$ (difference between the death and birth of a topological feature) denotes the \emph{lifetime} or \emph{persistence} of the topological feature. Parameter $p$ 
is varied to ignore or emphasize the persistent features, whereas the parameter $q$ is used to weight topological features that are born later at higher values of radius $r$. \\
\indent \textbf{Multi-Valued PSF:} Following a similar reasoning about multi-headed subnetworks as provided in the works of \cite{vaswani2017attention, veli_2018graph}, we too define the PSF via multiple pairs of values for $p$ and $q$, which are finally composed together. For this work, we have averaged over the PSF computed for different pairs of $p$ and $q$ in \eqref{eq:pers} as the composition function. This has twofold benefits: (i) multiple pairs for the tuple $(p, q)$ can provide different levels of \textit{penalty} and \textit{emphasis} on the topological features and (ii) we can bypass the manual effort of tuning PSF for a specific pair of $p$ and $q$. Henceforth, we refer to \emph{multi-valued PSF} as PSF for the remainder of this work. \\



\begin{figure*}[tbp]
	\centering
	\includegraphics[width=\linewidth,height=80mm]{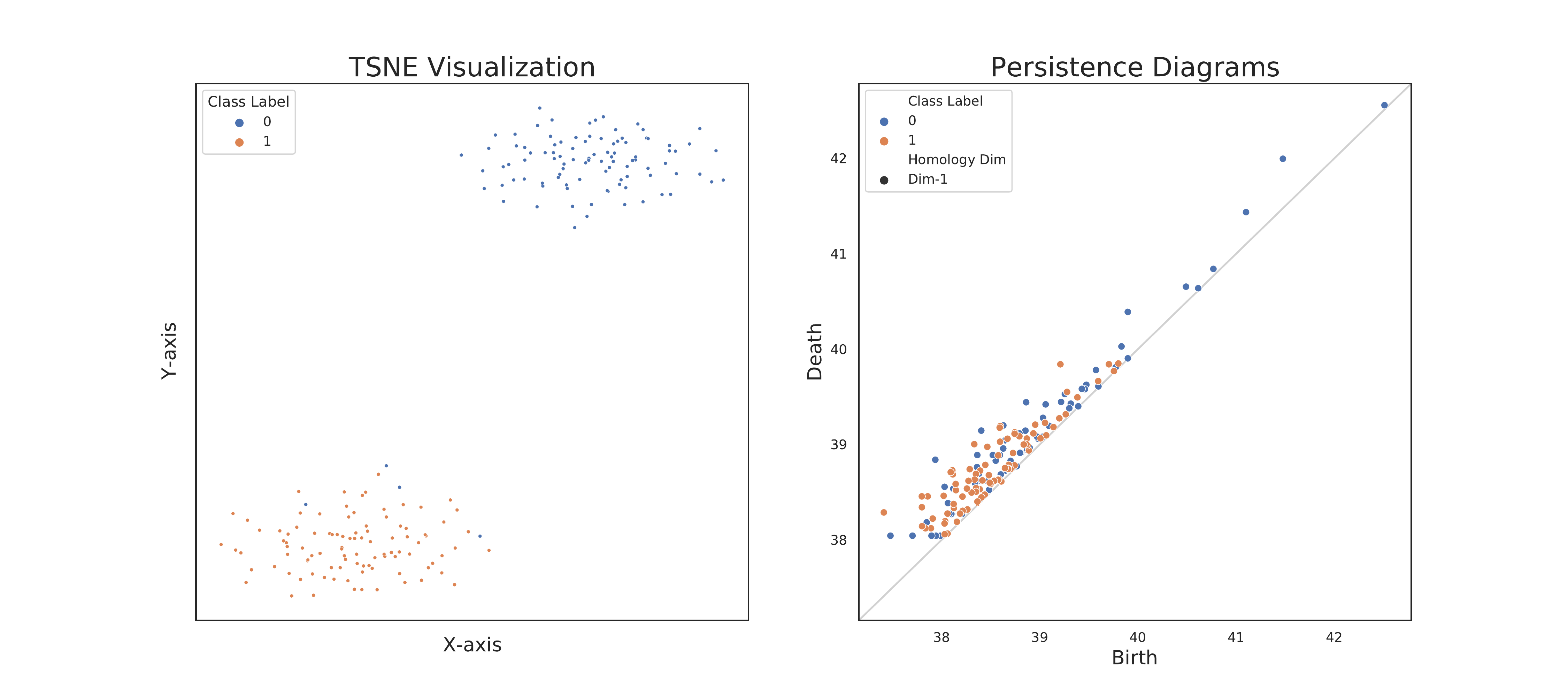}
	\caption{Sample TSNE and PD visualizations for SST-2 dataset}
	\label{fig:pd_sst}
\end{figure*}

\begin{figure*}[h]
	\begin{subfigure}[b]{0.33\textwidth}
		\centering
		\includegraphics[width=1.05\linewidth,height=55mm]{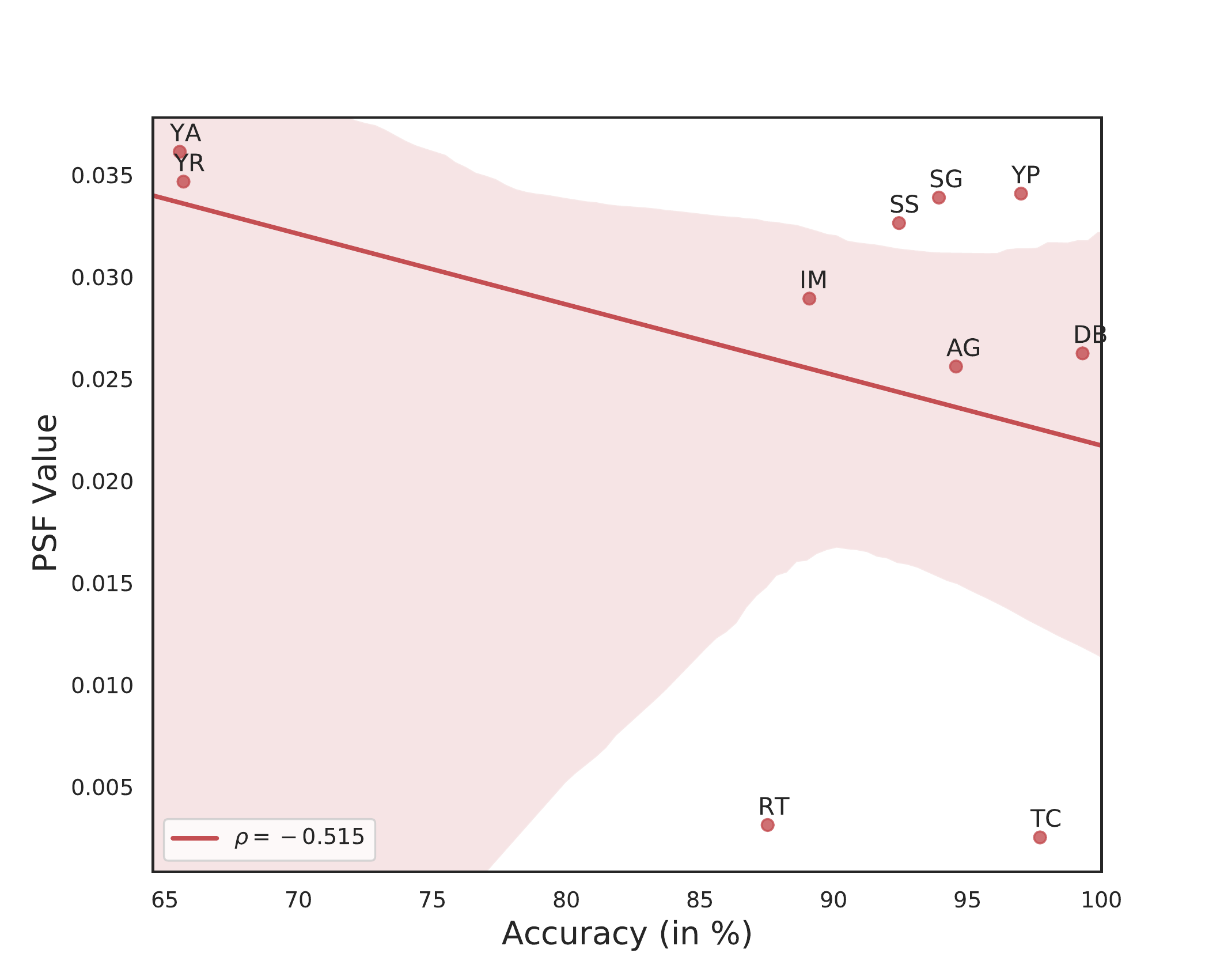}
		\caption{Accuracy vs PSF Value}
		\label{fig:acc_predictor_psf}
	\end{subfigure}%
	\begin{subfigure}[b]{0.33\textwidth}
		\centering
		\includegraphics[width=1.05\linewidth,height=55mm]{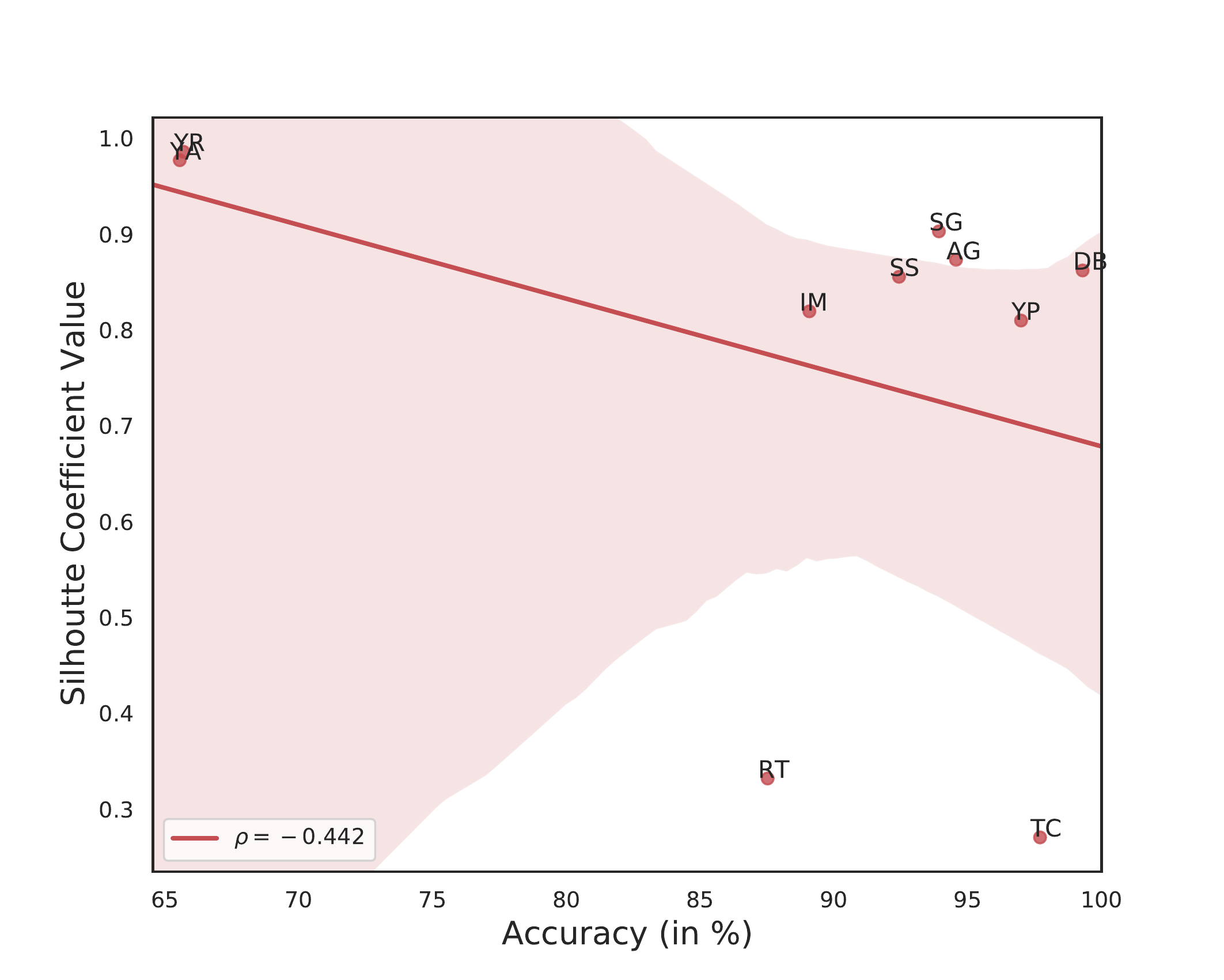}
		\caption{Accuracy vs Silhouette Coefficient}
		\label{fig:acc_predictor_sc}
	\end{subfigure}%
	\begin{subfigure}[b]{0.33\textwidth}
		\centering
		\includegraphics[width=1.05\linewidth,height=55mm]{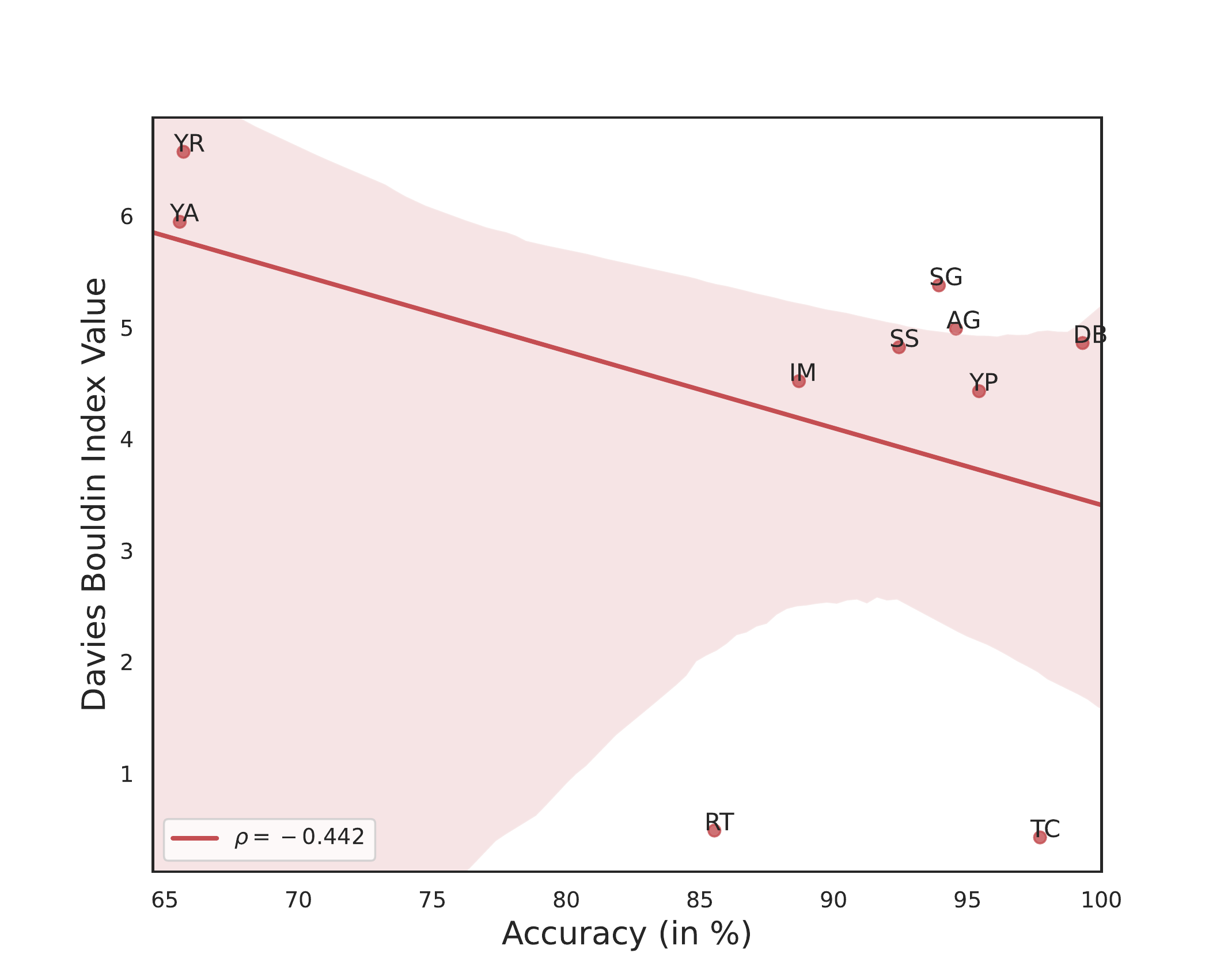}
		\caption{Accuracy vs Davies Bouldin Index}
		\label{fig:acc_predictor_dbi}
	\end{subfigure}
	
	\caption{Accuracy scores of the datasets against the estimator functions: PSF, Silhouette Coefficient and Davies Bouldin Index.}	
	\label{fig:acc_predictor}
\end{figure*}

\section{Empirical Study}
\subsection{Setup}
\label{inital-setup}
We validate the utility of PSF empirically using the BERT \cite{devlin-etal-2019-bert} model consisting of 12 layers and 12 attention heads per layer from \cite{wolf2020huggingfaces}. We consider ten widely used datasets, spanning small and large datasets alike, to show that PSF is agnostic to dataset size. Six of these are standard multi-class classification datasets: \textit{AG News}(AG), \textit{DBPedia 14}(DB), \textit{Sogou News}(SG), \textit{Yelp Review Full}(YR), \textit{Yahoo Answers Topics}(YA) and \textit{TREC}(TC), while the rest are sentiment polarity classification datasets: \textit{SST-2}(SS), \textit{IMDB}(IM), \textit{Yelp Polarity}(YP) and \textit{Rotten Tomatoes}(RT). The datasets are taken from the  \emph{huggingface}\footnote{https://huggingface.co/docs/datasets/}{datasets-library}. The experiments were performed on a single NVIDIA Titan-X GPU with $12$ GB CUDA memory.

For the PSF score outlined in \eqref{eq:pers}, we select multiple pairs of values for $p$ and $q$ from a very small set as: $\{p,q\} \in \mathcal{S} = \{2,3\} \times \{2,3\}$, rather than performing a naive grid search over a large set. We exclude values less than $2$ from the set because they provide weaker penalization and pairs with small persistence values can be considered as noise and ignored, while values greater than $3$ provide very strong penalization and diminish many persistence pairs. 

The construction of persistence diagrams is done using the well-known \textit{Vietoris-Rips filtration} from \emph{Ripser} \cite{ctralie2018ripser} library. To obtain the persistence diagrams $P_0(X)$ and $P_1(X)$, the \textbf{training samples} from each class of a given dataset are passed through the fine-tuned BERT and the corresponding vector representations of [CLS] token from the final layer are passed through the filtration mechanism. We compute the persistence diagrams upto homology group 1, ie, the holes/tunnels in the compact space.

\indent \textbf{Downsampling the Points:} Since the number of samples per class can be large depending upon the dataset used, we first use \textit{Kernel Density Estimation} (KDE) \cite{chen2017_kde} over the vector representations of the [CLS] tokens and then sample a desired number of points from KDE, which are finally passed through the filtration mechanism to compute the PDs. All the experiments have been performed over the points (of each class) sampled from KDE. Finally, the PSF score for a given dataset is computed by averaging the class-wise PSF scores.

In Figure~\ref{fig:pd_sst}, we juxtapose a t-SNE plot of BERT hidden states and the corresponding persistence diagram (only dim-1 topological features, i.e., the "holes" in the space) for the SST-2 dataset. Observe that for class label $0$ (blue points), there is a wider spread of birth-death pairs, where later birth indicates the formation of very large holes due to points covering a wider expanse. In comparison, points for class label $1$ (orange points) are more densely packed and have no points in the other cluster, therefore they cover a shorter range of birth in the persistence diagram.

	
	

	

\subsection{Results}
\label{subsec:results_}


\subsubsection{Estimating Test Accuracy via PSF}
\label{exp:accuracy_estimator}
In this section, we report the results of PSF as an estimator of the test performance of fine-tuned BERT models. This is done by calculating the \emph{Spearman rank correlation} $\rho$ of PSF to the test set accuracy of the corresponding datasets. It is noteworthy that in this experiment, we desire a \textit{negative correlation} because of the fact that the model performances are highly associated with the cluster quality of the vector representations in the latent space, i.e., more compact clusters result in lower PSF values (due to lower persistence in topology features) but result in higher accuracy due to improved class separation. Improved model performance with more compact clusters of data points has been previously shown by exhaustive experiments of \cite{DBLP:journals/corr/abs-1806-02679}. 

As baselines, we compare to the popular \textit{Silhouette coefficient} (SC) \cite{rousseeuw_198753}, where we subtract the value of SC on each dataset from 1 to ensure consistency in experiments, and the \textit{Davies Bouldin index} (DBI) \cite{davies_bouldin_4766909}. Both these methods have been widely used in the literature to assess the quality of the learned representations of the data by machine learning models. For reference, we point the reader to the works by\cite{chauhan2020FEW, article_dbi}. Figure \ref{fig:acc_predictor} shows the dataset accuracies versus the PSF values of the different estimator functions. We can observe that PSF, with a stronger negative correlation of $-0.515$, outperforms both SC and DBI with a correlation of $-0.442$, thus affirming our claims. 

We mentioned the practical utility of such estimators in section \ref{sec:intro}. Apart from comparing the performance across datasets, we expect that PSF can also be quite useful in the scenarios where we want to analyze and compare the performance of variants of a single model trained on different random seeds to select the best performing ones, which we will explore in depth in our future works. \\

\begin{figure}[h]
		\includegraphics[width=\linewidth]{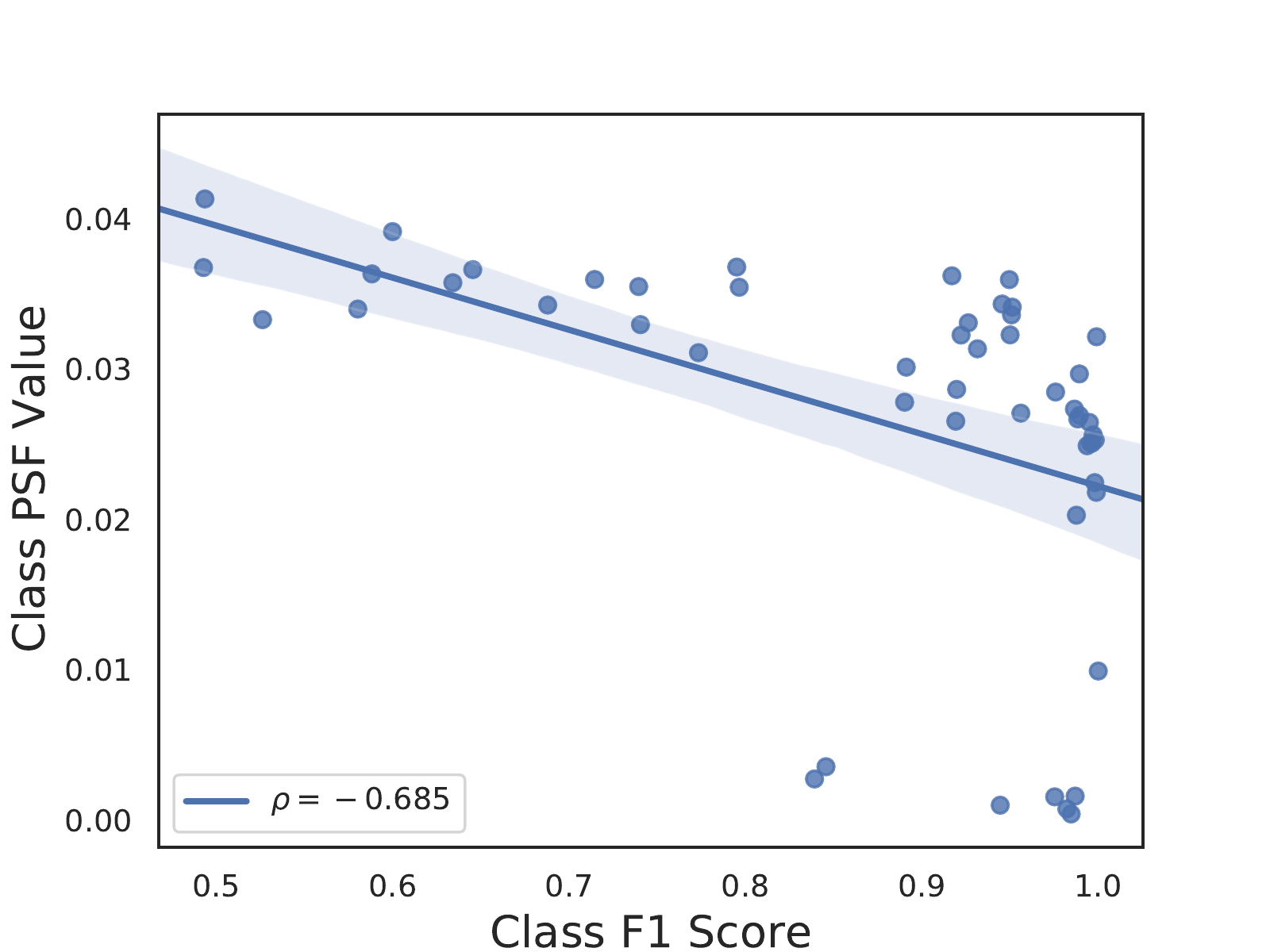}
	
	\caption{Classwise F1-scores vs the PSF values. 
	}	
	\label{fig:label_wise}
\end{figure}

\subsubsection{Fine-Grained Analysis}
We further conduct experiments to show that PSF can also be used for a more fine-grained analysis at a class-wise level and capture interesting properties about the same class labels across multiple datasets. This is done by calculating the Spearman correlation ($\rho$) of the f1-scores of the class labels to the corresponding class-wise PSF values. Similar to the previous experiment, we expect a strong negative correlation. Note that here we concatenate the PSF of the class labels across all the datasets in a single array (as opposed to averaging over an entire dataset). As shown in figure \ref{fig:label_wise}, PSF achieves a strong negative correlation of $-0.685$. 

From the qualitative viewpoint, we observed that the PSF value of the ``Business and Finance'' class was approximately the highest and its f1-score was approximately the lowest compared to other class labels across the datasets, while the PSF of ``Sports`` class was comparatively lower and its f1-score was higher than various other labels. 
We also observed that variance in the PSF values of the negative polarity class(label $0$) in sentiment datasets was high, whereas the positive polarity class (label $1$) had comparatively marginal variance, implying a distinct spread of the polarity classes across the datasets. We believe that such findings can help shed some light on the properties that the fine-tuned BERT models retain from the self-supervised pre-training phase pertaining to both data and tasks \cite{peters_2019_tune}. \\

\begin{figure}[tbp]
    \centering
    \includegraphics[width=\linewidth]{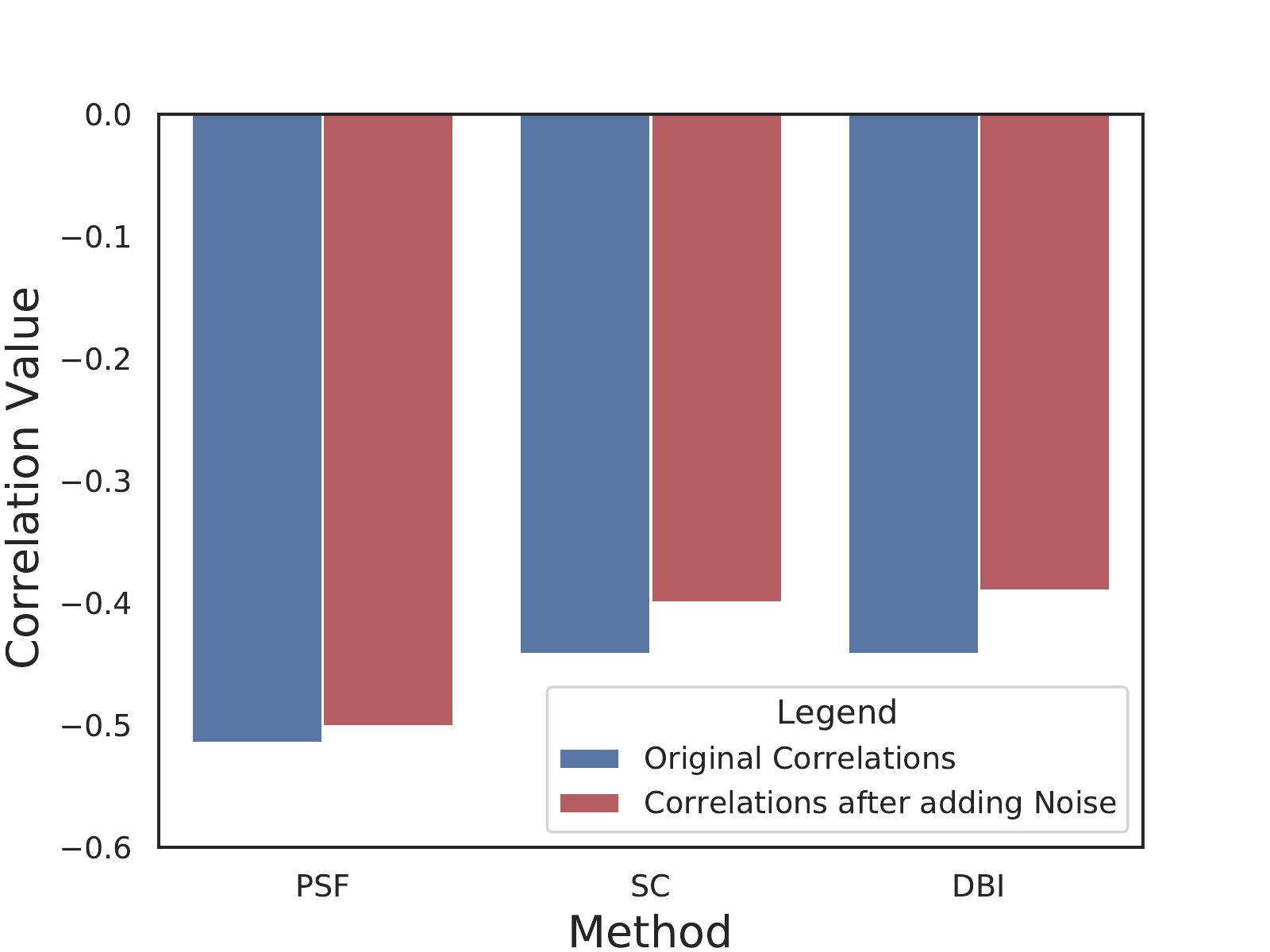}
	
	\caption{Stability analysis of PSF against SC and DBI.}	
	\label{fig:stability}
\end{figure}

\subsubsection{Stability of PSF}
Robustness against minor changes/perturbations in the inputs used to compute such estimators is one of the most important and desirable properties. Recent works such as \cite{zhang_2020BERTScore} in NLP and \cite{jiang2019predicting} in vision have focused on providing stable estimator and predictor metrics. Here, we show that PSF is robust to perturbations that are generated by adding random \textit{Gaussian noise} to the word embedding vectors of a subset of the input samples. Following the techniques outlined in\cite{zhang2018word}, we add noise $ \epsilon_h^i \sim \mathcal{N}(0,\,\sigma_h^{2})$ to each dimension $h$ of the $768$ dimensional embedding vectors of the words in input sample $i$ chosen for perturbation (here $\sigma_h^2$ is the variance of the corresponding dimension (column) of BERT's  embedding matrix). We then recompute the PSF scores (KDE step is also repeated) for the test accuracy estimation experiment in section \ref{exp:accuracy_estimator}. It is evident from the results shown in figure \ref{fig:stability} that the reduction in the correlation of PSF (reduction of $0.014$) is significantly lower than the reduction of $0.042$ and $0.052$ for baselines SC and DBI respectively, thus verifying the stability of PSF.

\subsection{PSF as an Estimator of Adversarial Vulnerability of trained Models} 
\label{sec:adv_vul}
Considering the experiments of the previous section as our basis, we present the core analysis results of our work in this section. 

It is well-known that deep natural language models are highly susceptible to adversarial attacks, as shown by various attack methods, elaborately described in   \cite{Zhang2019GeneratingTA}. Usually, the adversarial vulnerability is quantitatively evaluated by attacking the NLP model via multiple text samples and calculating the \textit{success rate} of the attacker, which, however, has various computational issues as described in section~\ref{sec:intro}. Thus, the question - ``Can we \textit{estimate} the success rate without actually performing the attack?'' becomes extremely pertinent and interesting. We explicitly point out that here we are trying to estimate the value of the success rate as an expectation over a large number of samples, thus providing a \emph{global perspective} of the attacker, rather than evaluating it per sample as done by \cite{corneanu_8953424}, which requires an actual input sample for evaluation, a bottleneck we try to overcome here. The following experiments show that PSF can serve as a good estimator of the success rates of both the black-box and white-box categories of attackers and substantially outperform the baselines as well. \\

\subsubsection{Setup}
The experimental setup for computing PSF as well as the datasets are retained from section \ref{inital-setup}. Here, we first describe the baselines used for comparison, then briefly mention the adversarial attack methods and lastly the setup for performing the attack.

\textbf{Baseline 1:} We call the first baseline as \textit{Adjusted Local Intrinsic Dimensionality (ALID)}, a modified version of Local Intrinsic Dimensionality (LID) proposed by \cite{ma2018characterizing}, originally used for the characterization of adversarial regions of neural networks. For our work, we use it for the characterization of the compactness of representations of the samples for each class label. For a given vector $\mathbf{x^c_j}$ in class $c$, we first calculate its ALID as follows:
\begin{equation}
 ALID(\mathbf{x^c_j}) = r_k(\mathbf{x^c_j}) \left(- \frac{1}{k} \sum_{i=1}^{k} log\frac{r_i(\mathbf{x^c_j})}{r_k(\mathbf{x^c_j})} \right)
\end{equation}
where $r_i(\mathbf{x^c_j})$ is the distance between $\mathbf{x^c_j}$ and its $i^{th}$ nearest neighbor within the samples of points of class $c$ and $r_k(\mathbf{x^c_j})$ is the maximum distance. Here, $k$ is the nearest neighbor hyperparameter (tuned manually). The ALID score for class $c$ is then obtained as
\begin{equation}
  ALID^c(.) = \frac{\frac{1}{N_c} \sum_{j=1}^{N_c} ALID(\mathbf{x^c_j})}{ \max \{ALID(\mathbf{x^c_j}) \mid j \in \{1,...,N_c\}\} } 
\end{equation}
where $N_c$ is the number of samples in class $c$. Finally, the ALID score for a given dataset is the average of class-wise ALID scores. 

\indent \textbf{Baseline 2:} We refer to the second baseline as the \textit{Adjusted Mahalanobis Score (AMS)}, which is a modification of the \textit{Mahalanobis distance-based confidence score} proposed by \cite{mad_lee}, originally used for detection of out-of-distribution and adversarial samples (generated by attack methods), both done simultaneously. Here, we use it as a Gaussian density-based scoring mechanism to measure the compactness of representations of the samples for each class label. For a given class $c$, we first calculate the empirical mean ($\hat{\mu}$) and covariance ($\hat{\mathbf{V}}$) as follows
\begin{equation}
 \hat{\mu} = \frac{1}{N_c} \sum_{i=1}^{N_c} \mathbf{x^c_i} , \\ \hat{\mathbf{V}} = \frac{1}{N_c} \sum_{i=1}^{N_c} (\mathbf{x^c_i} - \hat{\mu})(\mathbf{x^c_i} - \hat{\mu})^T
\end{equation}
where $N_c$ is the number of samples in class $c$. The AMS score for vector $\mathbf{x_j}$ in class $c$ is then calculated as
\begin{equation}
 AMS(\mathbf{x^c_j}) = (\mathbf{x^c_j} - \hat{\mu})^T \hat{\mathbf{V}}^{-1} (\mathbf{x^c_j} - \hat{\mu})
\end{equation}
Finally, the AMS score for class $c$ is obtained as:
\begin{equation}
 AMS^c(.) = \frac{\frac{1}{N_c} \sum_{j=1}^{N_c} AMS(\mathbf{x^c_j})}{\max \{AMS(\mathbf{x^c_j}) \mid j \in \{1,...,N_c\}\}}
\end{equation}
The AMS score for a given dataset is the average of class-wise AMS scores.

It is important to note that there do not exist any baselines for direct comparison in this experiment, thus we select the above two state-of-the-art methods from the literature on adversarial attacks on models.

\indent \textbf{Attack Methods:} We consider six state-of-the-art attack methods, including both \textit{black} and \textit{white} box attacks. These are: Probability Weighted Word Saliency (PWWS) \cite{pwws_ren_generating}, Genetic Algorithm (GA) \cite{alzantot_genetic}, TextFooler (TF) \cite{jin2020bert}, Textbugger Black-box (TB-B) as well as Textbugger White-box (TB-W) \cite{Li_2019} and lastly, Universal Adversarial Triggers (UAT) \cite{uat_wallace}.

Following \cite{jin2020bert}, we randomly select 1000 samples from the \textit{test} set of each dataset and perform adversarial attack via each of the six methods. Note that in this experiment, we desire a \textit{positive} correlation, since higher function values for PSF and both the baselines denote a larger spread in the hidden representation space and a higher probability of having many points near the non-linear decision boundaries, making the attacks easier \cite{karimi2020_decison}. \\

\begin{figure*}[h]

	\begin{subfigure}[b]{0.33\textwidth}
		\centering
		\includegraphics[width=\linewidth]{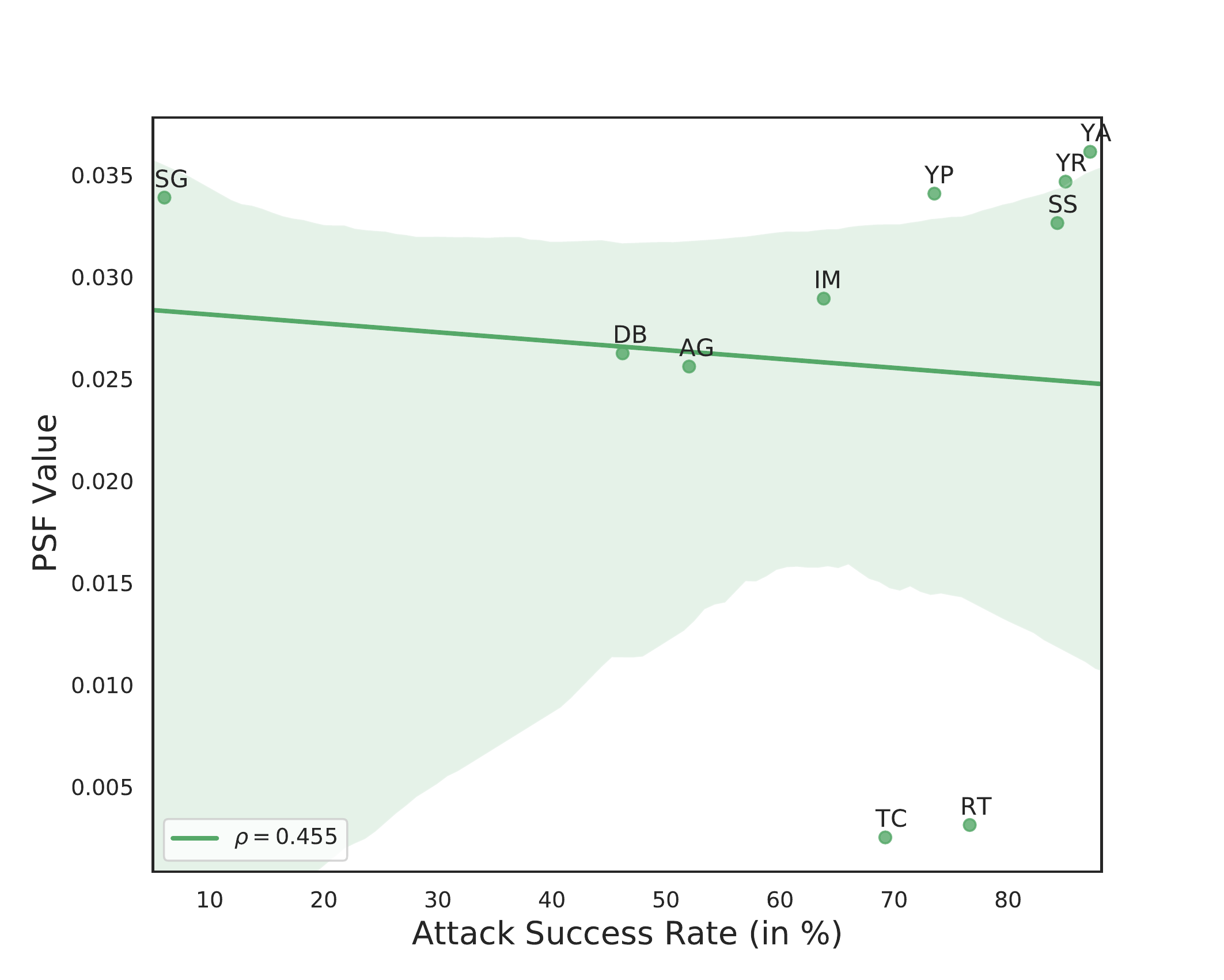}
		\caption{PWWS Attacker}
		\label{fig:adv_corrs_pwws}
	\end{subfigure}%
	\begin{subfigure}[b]{0.33\textwidth}
		\centering
		\includegraphics[width=\linewidth]{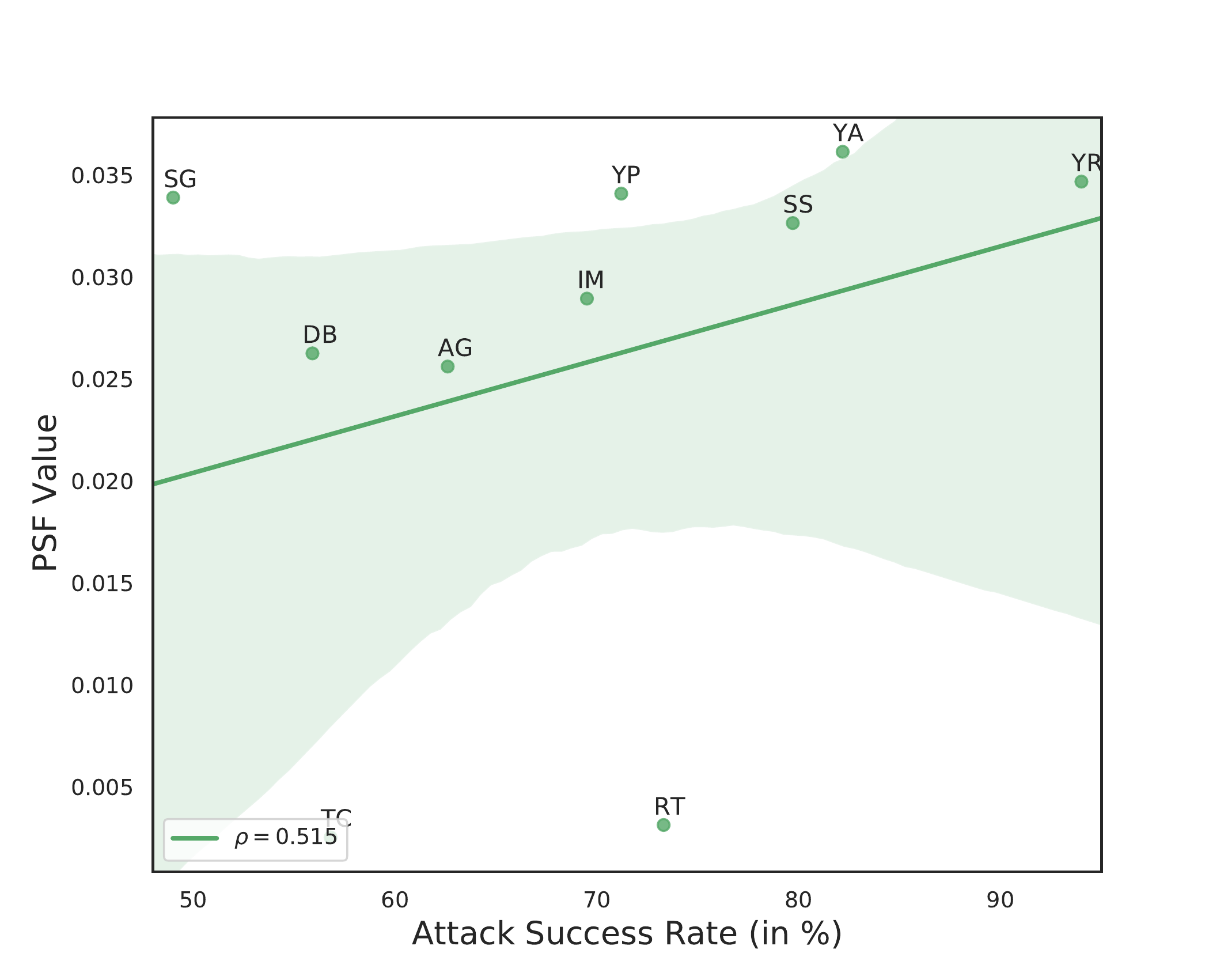}
		\caption{Textbugger Black-box Attacker}
		\label{fig:adv_corrs_tbb}
	\end{subfigure}%
	\begin{subfigure}[b]{0.33\textwidth}
		\centering
		\includegraphics[width=\linewidth]{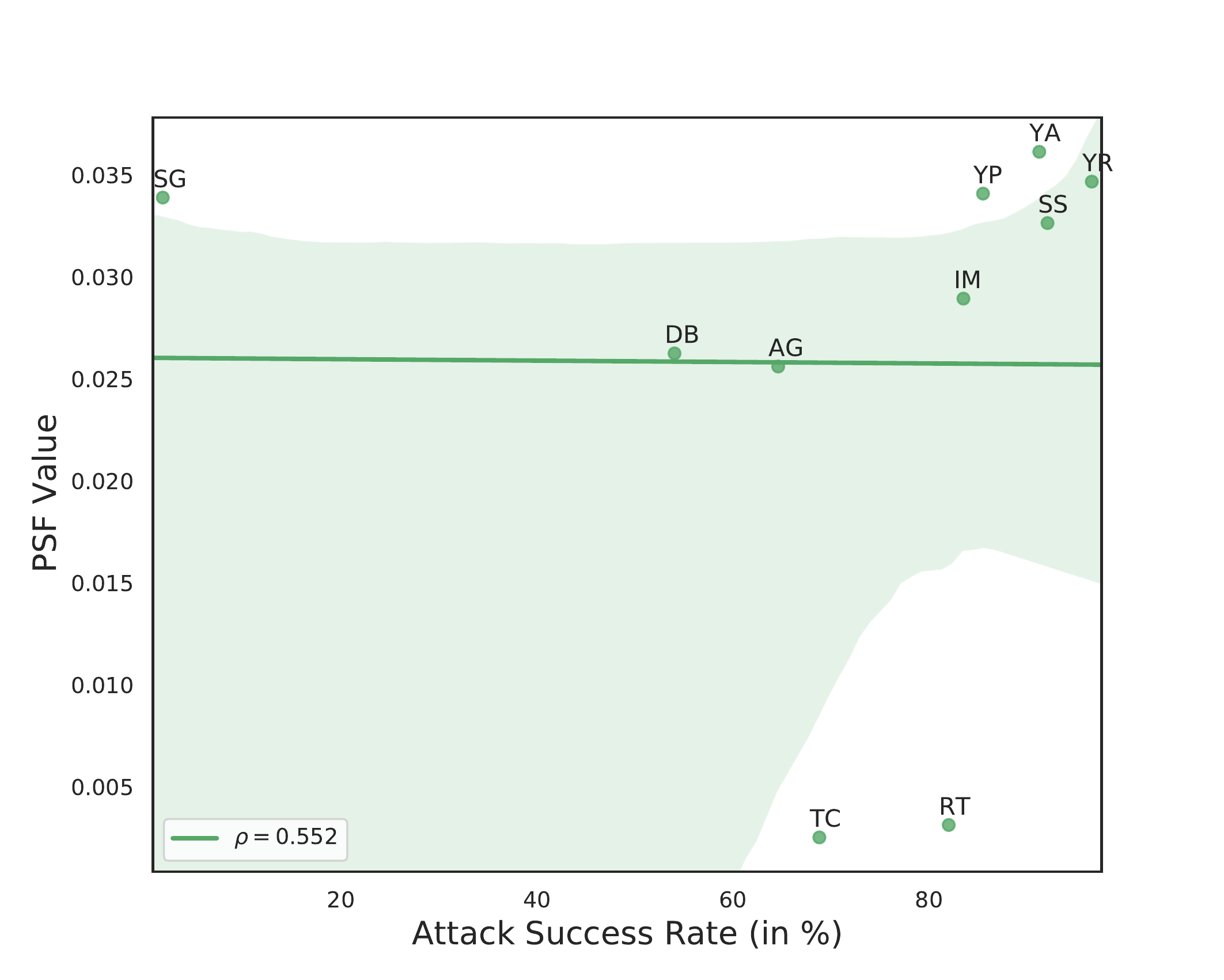}
		\caption{TextFooler Attacker}
		\label{fig:adv_corrs_tf}
	\end{subfigure}%
	
	\begin{subfigure}[b]{0.33\textwidth}
		\centering
		\includegraphics[width=\linewidth]{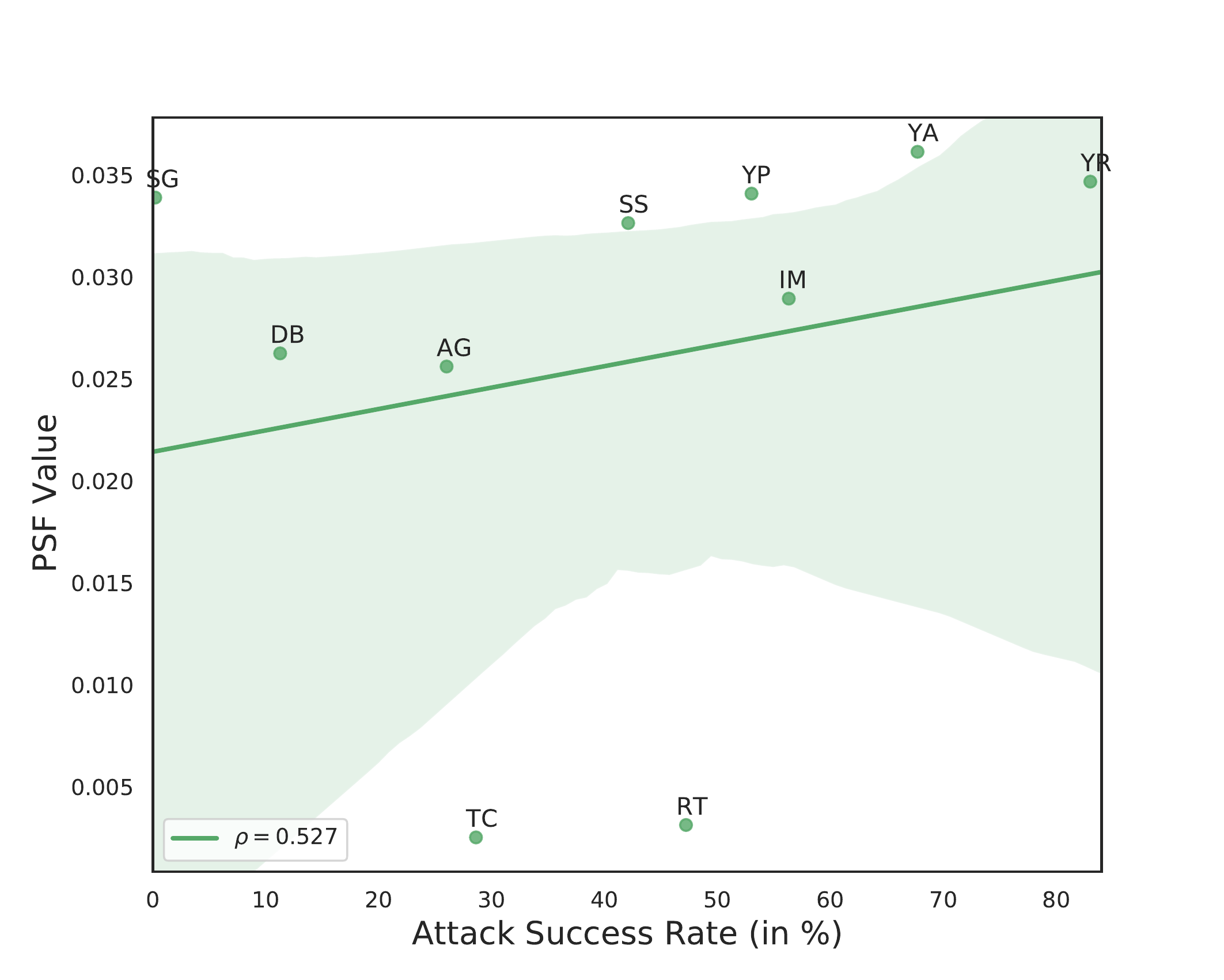}
		\caption{Genetic Attacker}
		\label{fig:adv_corrs_genetic}
	\end{subfigure}%
	\begin{subfigure}[b]{0.33\textwidth}
		\centering
		\includegraphics[width=\linewidth]{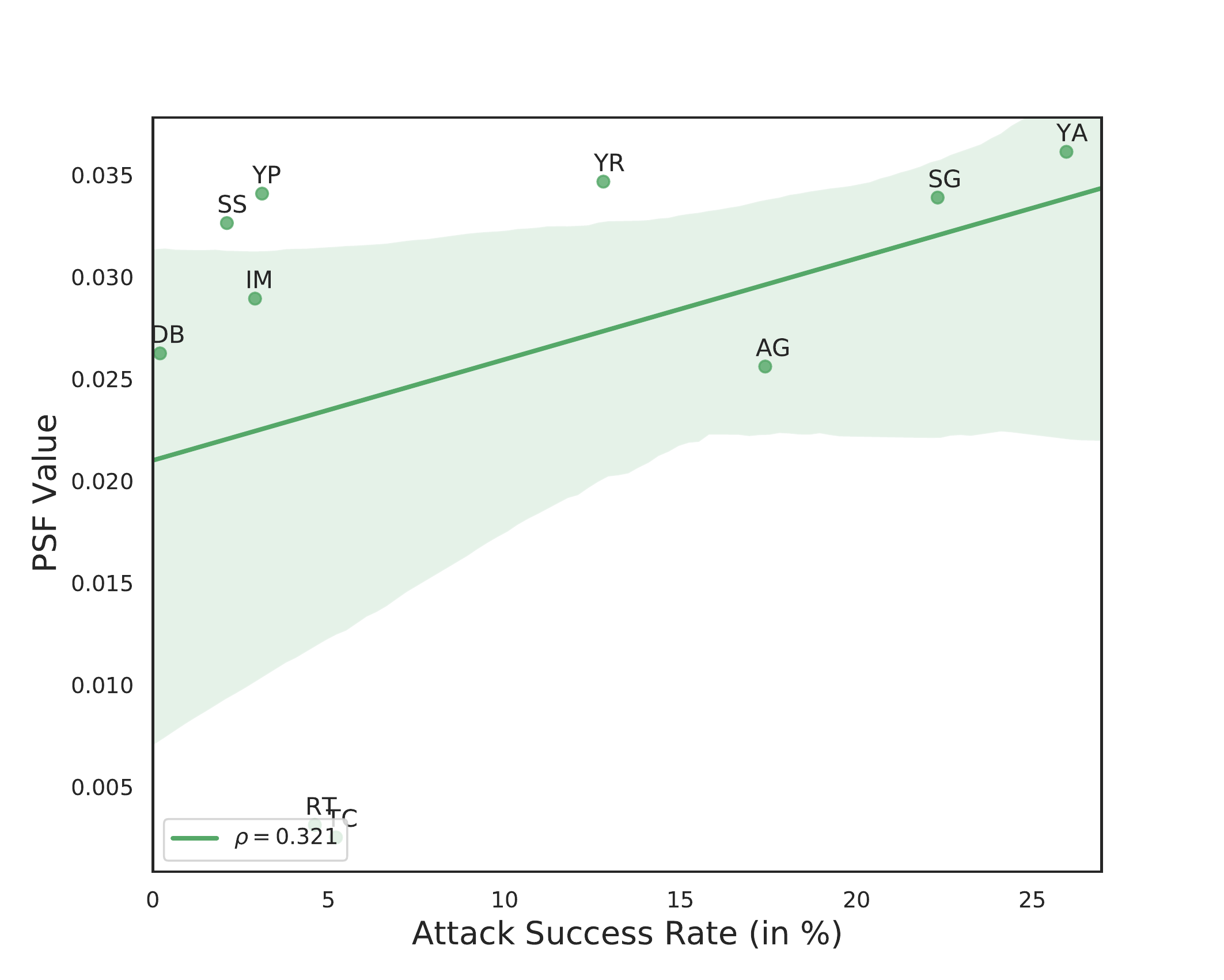}
		\caption{UAT Attacker}
		\label{fig:adv_corrs_uat}
	\end{subfigure}%
	\begin{subfigure}[b]{0.33\textwidth}
		\centering
		\includegraphics[width=\linewidth]{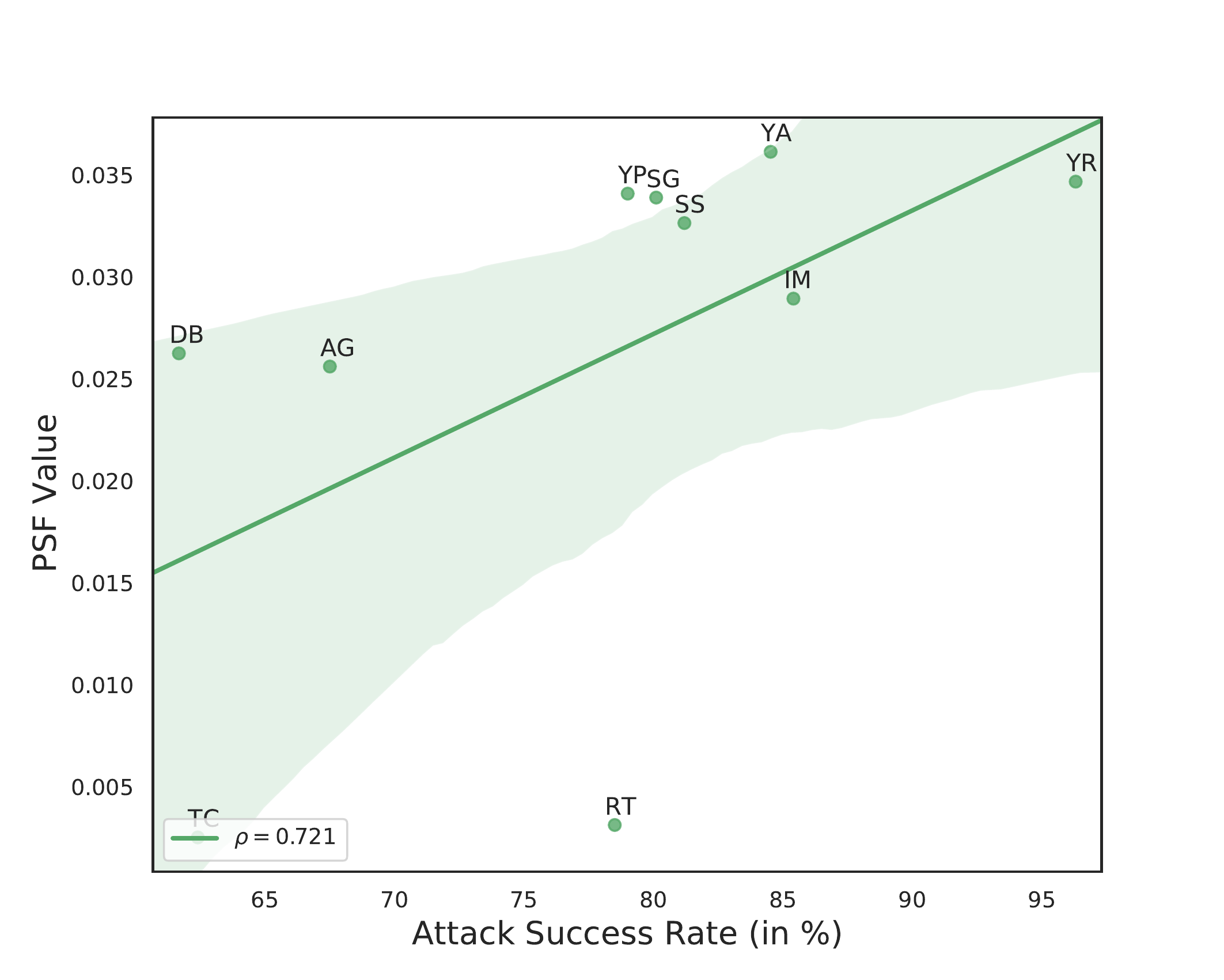}
		\caption{Textbugger White-box Attacker}
		\label{fig:adv_corrs_tbw}
	\end{subfigure}%
	
	\caption{Each plots shows the corresponding attack method's Success Rates (in \%) vs the PSF values on the $10$ datasets. Full forms of dataset abbreviations are same as in section \ref{inital-setup} . }	
	\label{fig:adv_corrs}
\end{figure*}

\begin{table*}[!htb]
\centering
\caption{Spearman correlations of PSF, ALID and AMS (row-wise) against the attack methods (column-wise). Best results in \textbf{bold}, second best \underline{underlined}. \% Improvements of PSF against each baseline are also reported.}
\label{table:adv_baselines}
\begin{tabular}{|c|c|c|c|c|c|c|}
\hline 
 & PWWS & TB-B & TF & GA & UAT & TB-W  \\
\hline
PSF (Ours) & \underline{0.4545} & $\bm{0.5151}$ & $\bm{0.5515}$ & $\bm{0.5272}$ & 0.3212 & $\bm{0.7212}$             \\
\hline

ALID   & 0.3696 & 0.1515 & 0.0420 & 0.0780 & $\bm{0.5393}$ & -0.0900      \\
\% Improvement of PSF over ALID & \emph{22.97\%} & \emph{240\%}  & \emph{1213\%} & \emph{575\%} & \emph{-40\%}  & \emph{901\%} \\
\hline

AMS  & $\bm{0.5030}$ & \underline{0.4060} & \underline{0.4787} & \underline{0.2727} & \underline{0.4181} & \underline{0.4545} \\
\% Improvement of PSF over AMS & \emph{-0.09\%} & \emph{26.8\%}  & \emph{15.2\%} & \emph{93.3\%} & \emph{-23.4\%}  & \emph{58.67\%} \\ 
\hline
\end{tabular}

\end{table*}

\subsubsection{Discussion}
Figure \ref{fig:adv_corrs} shows the correlation plots of PSF for each of the attack methods. From the results demonstrating the comparison of PSF to the baselines shown in Table \ref{table:adv_baselines}, it is evident that PSF outperforms both ALID and AMS by large margins for most of the attack methods. The \% improvements of PSF are extremely large as compared to ALID, wherein the improvements are more than $200\%$ on four attack methods, including both black-box and white-box. Against AMS, the improvements are more than $25\%$ on three attack methods, while the reductions on PWWS are negligible. UAT attacker seems to be an exception, where ALID outperforms both PSF and AMS. 

We can conclude from these results that PSF can serve as a good estimator of the attack success rates and is able to generalize well across attack methods from different categories measured on diverse datasets. Apart from estimating the vulnerabilities at the dataset level, we expect that such estimators can help us identify, from a set of multiple models trained on the same dataset, the least vulnerable ones, along similar lines as mentioned in Section \ref{exp:accuracy_estimator}. 

Note that it is non-trivial to provide theoretical claims and layout the conditions under which a given method will show a good correlation against an attacker due to the discrete nature of the adversarial attack task in NLP and variation in attack strategies, for eg: UAT's principal attack mechanism is different from standard white-box attacks. The same holds true for the genetic algorithm as compared to the greedy blank-out mechanism of Textbugger, and will be our focus in future works. 

\section{Conclusion and Future Work}
In this work, we proposed a novel scoring function named \textit{persistent scoring function (PSF)} using persistent homology to study the topological features present in BERT's hidden representation. Through exhaustive empirical studies, we showed that PSF can capture the homology of hidden state representations of BERT quite accurately and can be utilized as an estimator of the test set accuracies over a diverse range of classification datasets. PSF also provides some qualitative properties on a more fine-grained ``per class'' level and more importantly, it is more stable w.r.t to perturbations in the input space, thus realizing its practical utility. Lastly, we showed that PSF can also be used as an estimator of adversarial success rates generalizing across different categories of attack methods. PSF also has the advantage that it requires a few lines of post-processing code and can be utilized with any representation learning model.\\ 
\indent We hope that our work will pave a new and interesting direction for the community to realize the power of topological feature analysis to: (i) explain and interpret large-scale transformers variants such as BERT, RoBERTa, ALBERT, T5 etc and (ii) propose robust estimators and predictors for more diverse NLP tasks ranging from machine translation, dialogue evaluation to summarization in the future.

\bibliographystyle{IEEEtran}
\bibliography{paper}

\begin{thebibliography}{10}
\providecommand{\url}[1]{#1}
\csname url@samestyle\endcsname
\providecommand{\newblock}{\relax}
\providecommand{\bibinfo}[2]{#2}
\providecommand{\BIBentrySTDinterwordspacing}{\spaceskip=0pt\relax}
\providecommand{\BIBentryALTinterwordstretchfactor}{4}
\providecommand{\BIBentryALTinterwordspacing}{\spaceskip=\fontdimen2\font plus
\BIBentryALTinterwordstretchfactor\fontdimen3\font minus
  \fontdimen4\font\relax}
\providecommand{\BIBforeignlanguage}[2]{{%
\expandafter\ifx\csname l@#1\endcsname\relax
\typeout{** WARNING: IEEEtran.bst: No hyphenation pattern has been}%
\typeout{** loaded for the language `#1'. Using the pattern for}%
\typeout{** the default language instead.}%
\else
\language=\csname l@#1\endcsname
\fi
#2}}
\providecommand{\BIBdecl}{\relax}
\BIBdecl

\bibitem{vaswani2017attention}
A.~Vaswani, N.~Shazeer, N.~Parmar, J.~Uszkoreit, L.~Jones, A.~N. Gomez,
  L.~Kaiser, and I.~Polosukhin, ``Attention is all you need,'' \emph{arXiv
  preprint arXiv:1706.03762}, 2017.

\bibitem{devlin-etal-2019-bert}
J.~Devlin, M.-W. Chang, K.~Lee, and K.~Toutanova, ``{BERT}: Pre-training of
  deep bidirectional transformers for language understanding,'' in
  \emph{Proceedings of the 2019 Conference of the North {A}merican Chapter of
  the Association for Computational Linguistics}.

\bibitem{clark-etal-2019-bert}
K.~Clark, U.~Khandelwal, O.~Levy, and C.~D. Manning, ``What does {BERT} look
  at? an analysis of {BERT}{'}s attention,'' in \emph{Proceedings of the 2019
  ACL Workshop BlackboxNLP: Analyzing and Interpreting Neural Networks for
  NLP}.

\bibitem{hao2019visualizing}
Y.~Hao, L.~Dong, F.~Wei, and K.~Xu, ``Visualizing and understanding the
  effectiveness of bert,'' \emph{arXiv preprint arXiv:1908.05620}, 2019.

\bibitem{abnarquantifying}
S.~Abnar and W.~Zuidema, ``Quantifying attention flow in transformers,'' in
  \emph{Proceedings of the 58th Annual Meeting of the Association for
  Computational Linguistics}, Online, Jul. 2020.

\bibitem{nEURIPS2019_159c1ffe}
E.~Reif, A.~Yuan, M.~Wattenberg, F.~B. Viegas, A.~Coenen, A.~Pearce, and
  B.~Kim, ``Visualizing and measuring the geometry of bert,'' in \emph{Advances
  in Neural Information Processing Systems}, vol.~32, 2019.

\bibitem{schmidt2020bert}
F.~Schmidt and T.~Hofmann, ``Bert as a teacher: Contextual embeddings for
  sequence-level reward,'' 2020.

\bibitem{elsahar_galle_annotate}
H.~Elsahar and M.~Gall{\'e}, ``To annotate or not? predicting performance drop
  under domain shift,'' in \emph{Proceedings of the 2019 Conference on
  Empirical Methods in Natural Language Processing and the 9th International
  Joint Conference on Natural Language Processing (EMNLP-IJCNLP)}, Nov.

\bibitem{xia_2020_predicting}
M.~Xia, A.~Anastasopoulos, R.~Xu, Y.~Yang, and G.~Neubig, ``Predicting
  performance for natural language processing tasks,'' in \emph{Proceedings of
  the 58th Annual Meeting of the Association for Computational Linguistics}.

\bibitem{wang2020superglue}
A.~Wang, Y.~Pruksachatkun, N.~Nangia, A.~Singh, J.~Michael, F.~Hill, O.~Levy,
  and S.~R. Bowman, ``Superglue: A stickier benchmark for general-purpose
  language understanding systems,'' 2020.

\bibitem{rajpurkar2016squad}
P.~Rajpurkar, J.~Zhang, K.~Lopyrev, and P.~Liang, ``Squad: 100,000+ questions
  for machine comprehension of text,'' 2016.

\bibitem{sun_2017_ICCV}
C.~Sun, A.~Shrivastava, S.~Singh, and A.~Gupta, ``Revisiting unreasonable
  effectiveness of data in deep learning era,'' in \emph{Proceedings of the
  IEEE International Conference on Computer Vision (ICCV)}, Oct 2017.

\bibitem{effectiveness_4804817}
A.~Halevy, P.~Norvig, and F.~Pereira, ``The unreasonable effectiveness of
  data,'' \emph{IEEE Intelligent Systems}, vol.~24, no.~2, pp. 8--12, 2009.

\bibitem{edelsbrunner2000topological}
H.~Edelsbrunner, D.~Letscher, and A.~Zomorodian, ``Topological persistence and
  simplification,'' in \emph{Proceedings 41st annual symposium on foundations
  of computer science}.\hskip 1em plus 0.5em minus 0.4em\relax IEEE, 2000, pp.
  454--463.

\bibitem{edelsbrunner2010computational}
H.~Edelsbrunner and J.~Harer, \emph{Computational topology: an
  introduction}.\hskip 1em plus 0.5em minus 0.4em\relax American Mathematical
  Soc., 2010.

\bibitem{gracia2014}
A.~Gracia, S.~Gonz{\'a}lez, V.~Robles, and E.~Menasalvas, ``A methodology to
  compare dimensionality reduction algorithms in terms of loss of quality,''
  \emph{Information Sciences}, vol. 270, pp. 1--27, 2014.

\bibitem{jiang2019predicting}
Y.~Jiang, D.~Krishnan, H.~Mobahi, and S.~Bengio, ``Predicting the
  generalization gap in deep networks with margin distributions,'' 2019.

\bibitem{DBLP:journals/corr/abs-0712-2638}
F.~Chazal and S.~Oudot, ``Towards persistence-based reconstruction in euclidean
  spaces,'' \emph{CoRR}, vol. abs/0712.2638, 2007.

\bibitem{dataset_ot}
D.~Alvarez-Melis and N.~Fusi, ``Geometric dataset distances via optimal
  transport,'' in \emph{Advances in Neural Information Processing Systems},
  vol.~33, 2020.

\bibitem{nEURIPS2019_faf02b23}
A.~May, J.~Zhang, T.~Dao, and C.~R\'{e}, ``On the downstream performance of
  compressed word embeddings,'' in \emph{Advances in Neural Information
  Processing Systems}, vol.~32, 2019.

\bibitem{rieck2018neural}
B.~Rieck, M.~Togninalli, C.~Bock, M.~Moor, M.~Horn, T.~Gumbsch, and
  K.~Borgwardt, ``Neural persistence: A complexity measure for deep neural
  networks using algebraic topology,'' in \emph{International Conference on
  Learning Representations}, 2019.

\bibitem{corneanu_9156398}
C.~A. Corneanu, S.~Escalera, and A.~M. Martinez, ``Computing the testing error
  without a testing set,'' in \emph{2020 IEEE/CVF Conference on Computer Vision
  and Pattern Recognition (CVPR)}, 2020.

\bibitem{corneanu_8953424}
C.~A. Corneanu, M.~Madadi, S.~Escalera, and A.~M. Martinez, ``What does it mean
  to learn in deep networks? and, how does one detect adversarial attacks?'' in
  \emph{2019 IEEE/CVF Conference on Computer Vision and Pattern Recognition
  (CVPR)}, 2019.

\bibitem{zhu2013persistent}
X.~Zhu, ``Persistent homology: an introduction and a new text representation
  for natural language processing,'' in \emph{Proceedings of the Twenty-Third
  international joint conference on Artificial Intelligence}, 2013.

\bibitem{doshi2018using}
P.~Doshi, ``Using topological data analysis for text classification,'' Ph.D.
  dissertation, The University of North Carolina at Charlotte, 2018.

\bibitem{kushnareva-etal-2021-artificial}
L.~Kushnareva, D.~Cherniavskii, V.~Mikhailov, E.~Artemova, S.~Barannikov,
  A.~Bernstein, I.~Piontkovskaya, D.~Piontkovski, and E.~Burnaev, ``Artificial
  text detection via examining the topology of attention maps,'' in
  \emph{Proceedings of the 2021 Conference on Empirical Methods in Natural
  Language Processing}, 2021.

\bibitem{gabrielsson20a}
R.~B. Gabrielsson, B.~J. Nelson, A.~Dwaraknath, and P.~Skraba, ``A topology
  layer for machine learning,'' in \emph{Proceedings of the Twenty Third
  International Conference on Artificial Intelligence and Statistics}, ser.
  Proceedings of Machine Learning Research.\hskip 1em plus 0.5em minus
  0.4em\relax PMLR, 26--28 Aug 2020.

\bibitem{veli_2018graph}
P.~Veličković, G.~Cucurull, A.~Casanova, A.~Romero, P.~Liò, and Y.~Bengio,
  ``Graph attention networks,'' in \emph{International Conference on Learning
  Representations}, 2018.

\bibitem{wolf2020huggingfaces}
T.~Wolf, L.~Debut, V.~Sanh, J.~Chaumond, C.~Delangue, A.~Moi, P.~Cistac,
  T.~Rault, R.~Louf, M.~Funtowicz, J.~Davison, S.~Shleifer, P.~von Platen,
  C.~Ma, Y.~Jernite, J.~Plu, C.~Xu, T.~L. Scao, S.~Gugger, M.~Drame, Q.~Lhoest,
  and A.~M. Rush, ``Huggingface's transformers: State-of-the-art natural
  language processing,'' 2020.

\bibitem{ctralie2018ripser}
C.~Tralie, N.~Saul, and R.~Bar-On, ``{Ripser.py}: A lean persistent homology
  library for python,'' \emph{The Journal of Open Source Software}, vol.~3,
  no.~29, p. 925, Sep 2018.

\bibitem{chen2017_kde}
Y.-C. Chen, ``A tutorial on kernel density estimation and recent advances,''
  2017.

\bibitem{DBLP:journals/corr/abs-1806-02679}
K.~Kamnitsas, D.~C. de~Castro, L.~L. Folgoc, I.~Walker, R.~Tanno, D.~Rueckert,
  B.~Glocker, A.~Criminisi, and A.~V. Nori, ``Semi-supervised learning via
  compact latent space clustering,'' \emph{CoRR}, vol. abs/1806.02679, 2018.

\bibitem{rousseeuw_198753}
P.~J. Rousseeuw, ``Silhouettes: A graphical aid to the interpretation and
  validation of cluster analysis,'' \emph{Journal of Computational and Applied
  Mathematics}, vol.~20, pp. 53--65, 1987.

\bibitem{davies_bouldin_4766909}
D.~L. Davies and D.~W. Bouldin, ``A cluster separation measure,'' \emph{IEEE
  Transactions on Pattern Analysis and Machine Intelligence}, vol. PAMI-1,
  no.~2, pp. 224--227, 1979.

\bibitem{chauhan2020FEW}
J.~Chauhan, D.~Nathani, and M.~Kaul, ``Few-shot learning on graphs via
  super-classes based on graph spectral measures,'' in \emph{International
  Conference on Learning Representations}, 2020.

\bibitem{article_dbi}
H.~Jahangir, H.~Tayarani, S.~Sadeghi~Gougheri, M.~Aliakbar~Golkar, A.~Ahmadian,
  and A.~Elkamel, ``Deep learning-based forecasting approach in smart grids
  with micro-clustering and bi-directional lstm network,'' \emph{IEEE
  Transactions on Industrial Electronics}, pp. 1--1, 2020.

\bibitem{peters_2019_tune}
M.~E. Peters, S.~Ruder, and N.~A. Smith, ``To tune or not to tune? adapting
  pretrained representations to diverse tasks,'' in \emph{Proceedings of the
  4th Workshop on Representation Learning for NLP (RepL4NLP-2019)}.

\bibitem{zhang_2020BERTScore}
T.~Zhang*, V.~Kishore*, F.~Wu*, K.~Q. Weinberger, and Y.~Artzi, ``Bertscore:
  Evaluating text generation with bert,'' in \emph{International Conference on
  Learning Representations}, 2020.

\bibitem{zhang2018word}
D.~Zhang and Z.~Yang, ``Word embedding perturbation for sentence
  classification,'' \emph{arXiv preprint arXiv:1804.08166}, 2018.

\bibitem{Zhang2019GeneratingTA}
W.~E. Zhang, Q.~Z. Sheng, and A.~A.~F. Alhazmi, ``Generating textual
  adversarial examples for deep learning models: A survey,'' \emph{ArXiv}, vol.
  abs/1901.06796, 2019.

\bibitem{ma2018characterizing}
X.~Ma, B.~Li, Y.~Wang, S.~M. Erfani, S.~Wijewickrema, G.~Schoenebeck, M.~E.
  Houle, D.~Song, and J.~Bailey, ``Characterizing adversarial subspaces using
  local intrinsic dimensionality,'' in \emph{International Conference on
  Learning Representations}, 2018.

\bibitem{mad_lee}
K.~Lee, K.~Lee, H.~Lee, and J.~Shin, ``A simple unified framework for detecting
  out-of-distribution samples and adversarial attacks,'' in \emph{Advances in
  Neural Information Processing Systems}, vol.~31, 2018.

\bibitem{pwws_ren_generating}
S.~Ren, Y.~Deng, K.~He, and W.~Che, ``Generating natural language adversarial
  examples through probability weighted word saliency,'' in \emph{Proceedings
  of the 57th Annual Meeting of the Association for Computational Linguistics}.

\bibitem{alzantot_genetic}
M.~Alzantot, Y.~Sharma, A.~Elgohary, B.-J. Ho, M.~Srivastava, and K.-W. Chang,
  ``Generating natural language adversarial examples,'' in \emph{Proceedings of
  the 2018 Conference on Empirical Methods in Natural Language Processing}.

\bibitem{jin2020bert}
D.~Jin, Z.~Jin, J.~T. Zhou, and P.~Szolovits, ``Is bert really robust? a strong
  baseline for natural language attack on text classification and entailment,''
  2020.

\bibitem{Li_2019}
\BIBentryALTinterwordspacing
J.~Li, S.~Ji, T.~Du, B.~Li, and T.~Wang, ``Textbugger: Generating adversarial
  text against real-world applications,'' \emph{Proceedings 2019 Network and
  Distributed System Security Symposium}, 2019. [Online]. Available:
  \url{http://dx.doi.org/10.14722/ndss.2019.23138}
\BIBentrySTDinterwordspacing

\bibitem{uat_wallace}
E.~Wallace, S.~Feng, N.~Kandpal, M.~Gardner, and S.~Singh, ``Universal
  adversarial triggers for attacking and analyzing {NLP},'' in
  \emph{Proceedings of the 2019 Conference on Empirical Methods in Natural
  Language Processing and the 9th International Joint Conference on Natural
  Language Processing (EMNLP-IJCNLP)}, 2019.

\bibitem{karimi2020_decison}
H.~Karimi, T.~Derr, and J.~Tang, ``Characterizing the decision boundary of deep
  neural networks,'' 2020.

\end{thebibliography}

\end{document}